\documentclass[lettersize,conference]{IEEEtran}
\usepackage{amsmath,amssymb,amsfonts}
\usepackage{utfsym}
\usepackage{algorithmic}
\usepackage{algorithm}
\usepackage{array}
\usepackage[caption=false,font=normalsize,labelfont=sf,textfont=sf]{subfig}
\usepackage{textcomp}
\usepackage{stfloats}
\usepackage{url}
\usepackage{verbatim}
\usepackage{graphicx}
\usepackage{cite}
\usepackage{booktabs}
\usepackage{amsthm}
\usepackage{color}
\usepackage{xcolor}
\usepackage{soul}
\usepackage{multirow}
\usepackage{boldline} 
\usepackage{threeparttable}
\usepackage{hyperref}

\begin{document}

\title{Can Deep Network Balance Copy-Move Forgery Detection and Distinguishment?}

\author{Shizhen Chang
\thanks{S. Chang is with the Institute of Advanced Research and Artificial Intelligence, AI4RS group, 1030 Vienna, Austria}}

\markboth{Journal of \LaTeX\ Class Files,~Vol.~14, No.~8, August~2021}%
{Shell \MakeLowercase{\textit{et al.}}: A Sample Article Using IEEEtran.cls for IEEE Journals}

\IEEEpubid{0000--0000/00\$00.00~\copyright~2021 IEEE}

\maketitle

\begin{abstract}
Copy-move forgery detection is a crucial research area within digital image forensics, as it focuses on identifying instances where objects in an image are duplicated and placed in different locations. The detection of such forgeries is particularly important in contexts where they can be exploited for malicious purposes. Recent years have witnessed an increased interest in distinguishing between the original and duplicated objects in copy-move forgeries, accompanied by the development of larger-scale datasets to facilitate this task. However, existing approaches to copy-move forgery detection and source/target differentiation often involve two separate steps or the design of individual end-to-end networks for each task. In this paper, we propose an innovative method that employs the transformer architecture in an end-to-end deep neural network. Our method aims to detect instances of copy-move forgery while simultaneously localizing the source and target regions. By utilizing this approach, we address the challenges posed by multi-object copy-move scenarios and report if there is a balance between the detection and differentiation tasks. To evaluate the performance of our proposed network, we conducted experiments on two publicly available copy-move datasets. The results and analysis aims to show the potential significance of our focus in balancing detection and distinguishment result and transferring the trained model in different datasets in the field.
\end{abstract}


\section{Introduction}

\IEEEPARstart{T}{he} exponential growth of digital images on the internet, fueled by the increasing diversity of social media platforms, has brought forth significant challenges. One of the prominent issues is the ease with which digital images can be manipulated using easily accessible tools, leading to a surge in incidents of image tampering. Among various techniques employed for image tampering, copy-move forgery \cite{christlein2012evaluation} stands out as one of the most commonly utilized and easily executed methods. This technique involves duplicating a specific region, known as the source region, within an image. Subsequently, the duplicated region is manipulated through processes such as scaling, rotating, or color adjustment before being pasted into another region, termed the target region, within the same image. Copy-move forgery serves as a means to conceal or duplicate objects within an image for malicious purposes \cite{birajdar2013digital}. Instances of copy-move forgery, such as the circulation of fake news containing manipulated images in political contexts, have the potential to confuse the public and contribute to political biases. Similarly, the malicious manipulation of evidence within legal proceedings or the falsification of experimental results in academic papers can have severe consequences, including judicial injustice and academic misconduct.

Hence, the development of image forensic methods for copy-move forgery detection is of paramount importance. In particular, distinguishing between the source and target regions within an image is crucial to accurately identify manipulated areas—a process commonly referred to as copy-move source/target distinguish (CMSTD). This technique finds relevance in numerous scenarios, including legal investigations, journalism, scientific research, and digital art forensics. In legal investigations, CMSTD serves the purpose of discerning original evidence from manipulated content. For instance, it can aid in determining whether a weapon has been added or removed from an image, thus assisting in establishing the authenticity of evidence. Similarly, in journalism, CMSTD plays a pivotal role in verifying the genuineness of images employed in news stories. Furthermore, CMSTD finds application in scientific research to identify tampered images in published papers. By doing so, it safeguards the reliability and integrity of research findings. Additionally, in the realm of digital art forensics, CMSTD aids in the detection of image manipulation within artistic works. This aspect becomes crucial for purposes such as copyright protection and plagiarism detection. Given the increasing prevalence of digital image tampering, CMSTD has emerged as a critical technique for ensuring the authenticity and dependability of digital images across various domains.

As the prevalence of copy-move forgery continues to increase, there is a growing demand for the development of accurate and efficient detection methods. In recent years, various techniques have been proposed to address this issue. Feature-based algorithms \cite{amerini2011sift, li2014segmentation} employ feature extraction techniques such as SIFT and segmentation to identify duplicated regions. Similarly, block-based algorithms \cite{cozzolino2015efficient, ardizzone2015copy} utilize block comparisons for detection purposes. However, distinguishing between the source and target regions remains a challenging task in copy-move forgery detection. The rapid advancements in deep learning within the field of computer vision \cite{he2016deep, xu2022ai, chang2023dsfer} have motivated numerous studies that leverage deep feature analysis for copy-move forgery detection. For instance, Rao et al. \cite{rao2016deep} proposed a convolutional neural network (CNN) that utilizes high-pass filters as the initialized layer to identify copy-move regions. Wu et al. \cite{wu2018image} developed an end-to-end deep neural network capable of extracting block features and analyzing self-correlation between feature pixels to identify similar regions in copy-move forgery images. Additionally, Zhong et al. \cite{zhong2019end} proposed a dense inception network incorporating pyramid feature extractors, correlation matching blocks, and hierarchical post-processing modules to accurately localize copy-move regions.

However, the previously mentioned methods only occupy in binary detection where the duplicated regions are highlighted and the pristine backgrounds are suppressed. In recent years, research aims to distinguish source or target those regions belongs to also comes to public attention. Wu et al. \cite{wu2018busternet} proposed the BusterNet which consists of a similarity detection branch and a 
 manipulation detection branch to detect and distinguish copy-move regions in the images. Later, Chen et al. \cite{chen2020serial} proposed a cascaded network based on the architecture of BusterNet, which contains a copy-move similarity detection network (CMSDNet) and a source/target region distinguishment network (STRDNet). Different from BusterNet, the source/target localization is learned from the detection map of the CMSDNet, therefore, this method is a two-step netnetwork and the two subnetworks should be trained separately. Islam et al. \cite{islam2020doa} proposed a dual-order attentive generative adversarial network (DOA-GAN) that adopts the dual-order attention module to extract location-aware features and the atrous spatial pyramid pooling blocks to extract global features. Besides, Barni et al. \cite{barni2020copy} designed a multi-branch CNN to differentiate the source and target regions between two nearly duplicated regions with a hypothesis testing framework given
the binary localization mask. Zhang et al. \cite{zhang2022cnn} proposed a CNN transformer based Generative adversarial network that first combines transformer for feature extraction in copy-move forgery localization. 
\begin{figure*}[!t]
\centering
\includegraphics[width=0.9\linewidth]{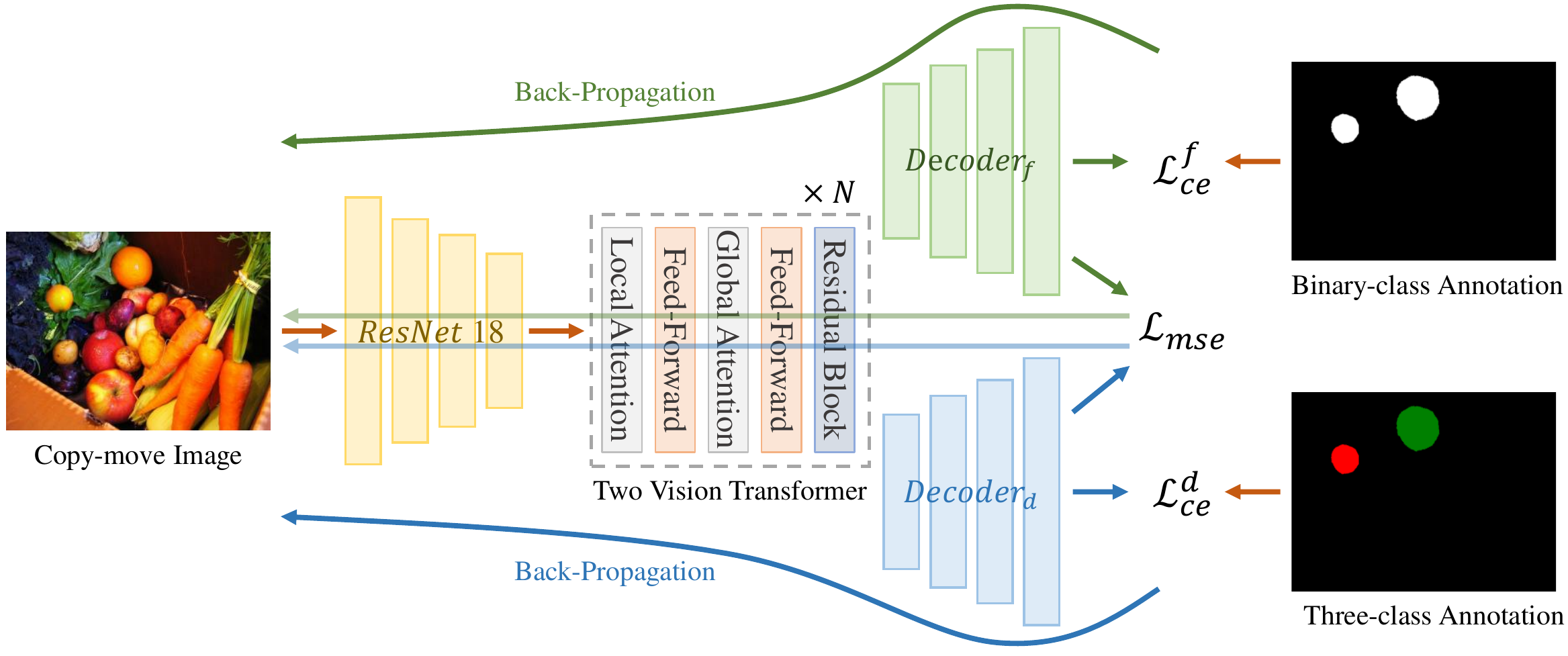}
\caption{The overall flowchart of the proposed network. The network consisting of three main components: a feature extractor, a transformer-based encoder, and a two-branch decoder. The ResNet18 is utilized as the backbone network. The extracted features are then passed through a transformer-based encoder, which is constructed using a combination of local attention and global attention mechanisms. The final stage involves the two-branch decoder, which is responsible for forgery detection and source/target distinguishment.}
\label{flowchart}
\end{figure*}

However, the detection and distinction of copy-move regions still pose significant challenges due to several factors. Firstly, the presence of substantial differences among various datasets makes it difficult to develop a method that can be effectively transferred across different scenarios. Secondly, accurately co-localizing the source and target regions as forged areas proves to be a complex task. Existing methods typically address these two tasks independently or rely on binary class detection obtained from a three-class segmentation mask. In this paper, we aim to investigate whether a CNN network can achieve a balanced performance in terms of forgery detection and source/target localization.

The remaining sections of the paper are organized as follows: Section II introduces the proposed network. In Section III, we present the experimental settings and provide an analysis of the results. Finally, Section IV summarizes the findings and concludes this study.

\section{Methodology}
\subsection{Overview}
The proposed research aims to evaluate the effectiveness of an end-to-end convolutional neural network in achieving a balanced performance between copy-move detection and localization. The network's ability to learn the co-relationship between source and target regions, instead of relying solely on memorizing their locations during training, is crucial due to variations in size and texture characteristics.

In Fig. \ref{flowchart}, the network architecture is illustrated. It consists of three main components: a feature extractor, a transformer-based encoder, and a two-branch decoder responsible for generating the detection map and source/target distinguishment map. The input image is assumed to have dimensions of $256\times256\times3$. Initially, the input image $X$ undergoes feature extraction using ResNet18 to extract deep features. Subsequently, a two-vision transformer is employed as the encoder to highlight the manipulated regions within the image feature map. Finally, the resulting features are fed into forgery detection and localization decoders, which share a similar architecture. These decoders upsample the feature maps to match the original image size, generating the binary-class detection map $\hat{Y}^f$ and the three-class segmentation map $\hat{Y}^d$, respectively.
\subsection{Transformer-based Encoder}
The transformer-based encoder leverages the power of two vision transformers to accurately localize copy-move regions within the deep feature. This aspect of the encoder plays a crucial role in capturing subtle and intricate modifications by computing the inner relationship among different feature patches. This is particularly challenging for convolutional neural networks that rely on direct memorization of related features from annotations. The design of the encoder takes inspiration from Twins\cite{chu2021twins}, a notable work in the field.

To illustrate the functioning of the encoder, refer to Fig. \ref{flowchart}. The local attention mechanism divides the features into multiple feature patches, enabling multi-head self-attention to be performed on each individual patch. This local attention mechanism helps in capturing fine-grained details and local dependencies within the feature space. On the other hand, the global attention mechanism conducts multi-head self-attention at a global level, allowing the encoder to capture broader contextual information and long-range dependencies. The enhanced deep features obtained from the attention mechanisms are further processed by a residual block \cite{chang2023changes} . This block comprises three 2D convolutional layers, with each layer followed by a Rectified Linear Unit (ReLU) activation function. The purpose of the residual block is to introduce non-linearity and learn more abstract representations that contribute to improved feature representations.

\subsection{Two Branch Decoder}

\begin{figure}[!t]
\centering
\includegraphics[width=0.6\linewidth]{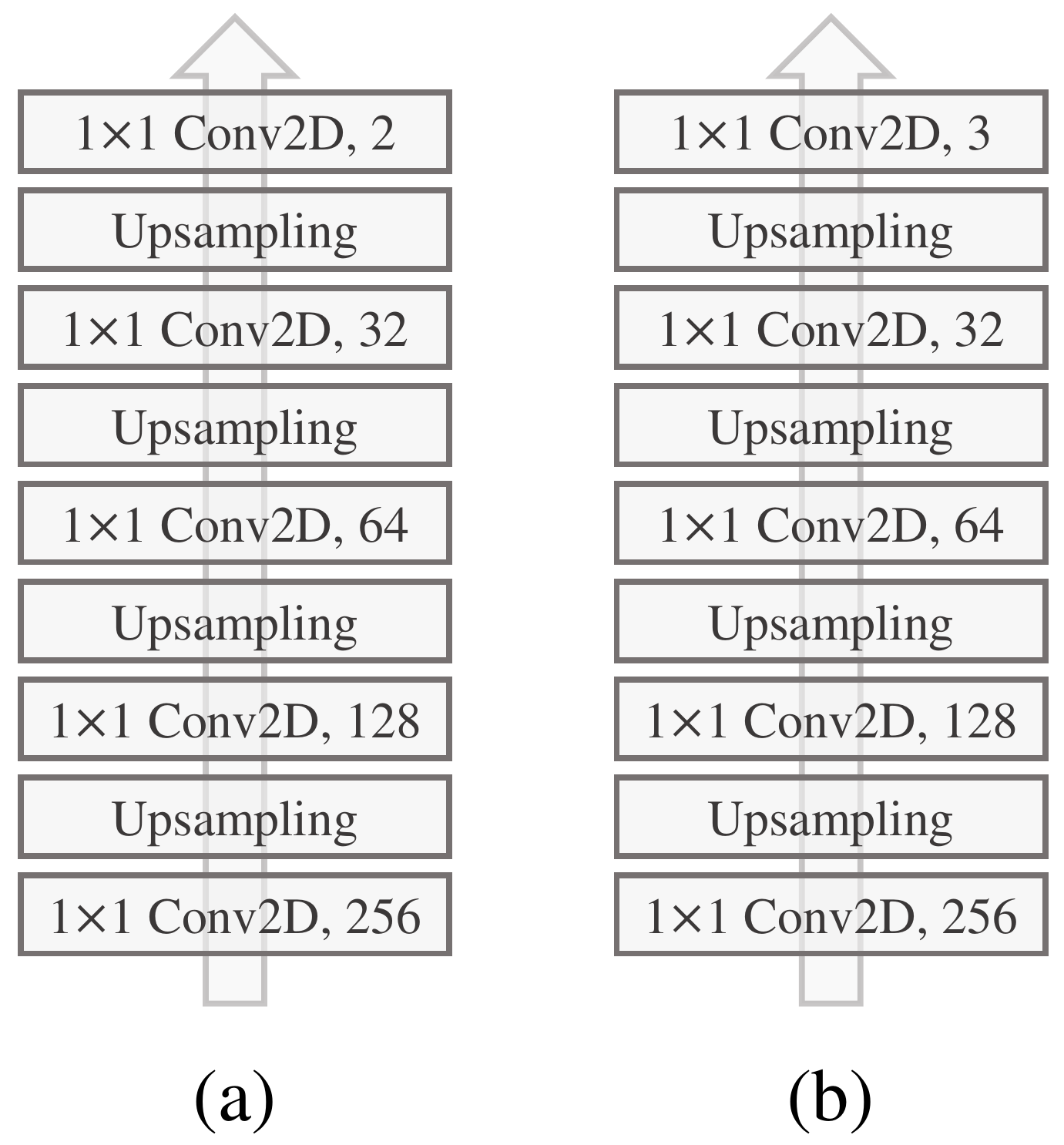}
\caption{The structure of the decoders. (a) $Decoder_f$ and (b) $Decoder_d$. }
\label{ft2}
\end{figure}

In order to effectively identify and delineate the copied regions within an image, our approach employs a dual-branch decoder architecture. This architecture comprises two distinct branches, namely the forgery detection branch and the source/target distinguishment branch. The decoder component itself is composed of four layers of 2D convolution, each of which is subsequently followed by an up-sampling layer.

Fig. \ref{ft2} illustrates the structure of the decoders, highlighting the differentiation between the forgery detection branch, denoted as $Decoder_f$, and the source/target distinguishment branch, denoted as $Decoder_d$. The differentiating factor lies in the final layer of each branch. Specifically, the forgery detection branch produces a binary-class map by generating a two-channel output, while the source/target distinguishment branch generates a three-channel output, resulting in a three-class map.

This design choice enables our system to accurately identify the areas where copying has occurred within the image. The binary-class map serves to indicate the presence or absence of copy-move regions, while the three-class map facilitates the differentiation between the source and target regions involved in the forgery. By utilizing these distinct decoder branches, our method aims to improve the precision and comprehensiveness of copy-move region detection.

The entire network is trained using a cross-entropy loss minimization approach for two branches, which involve comparing the predicted outputs with their corresponding annotations. The loss functions for each branch are defined as follows:
\begin{equation}
\begin{split}
&\mathcal{L}_{ce}^{f}=-\frac{1}{hw}\sum(Y_f \log\hat{Y}_f +(1-Y_f)\log(1-\hat{Y}_f)), \\
&\mathcal{L}_{ce}^{d}=-\frac{1}{hw}\sum_{c=1}^{3} (Y_d^c\log(\hat{Y}_d^c)),
\end{split}
\end{equation}
where $\hat{Y}_f$ and $\hat{Y}_d$ represent the predicted results of the $Decoder_f$ and $Decoder_d$, respectively. $Y_f$ and $Y_d$ denote the binary-class annotation map and three-class annotation map, respectively.

To ensure balanced performance between the two tasks, it is important for the predicted binary detection map and the three-class segmentation map to be consistent. To achieve this, the source and target classes of $\hat{Y}_d$ are combined into a single class. The mean squared error (MSE) between $\hat{Y}_f$ and $\hat{Y}_d$ is then calculated as follows:

\begin{equation}
\mathcal{L}_{mse}=\mathbb{E}(\hat{Y}_f; \hat{Y}_d)[(\hat{Y}_f -\hat{Y}_d)^2].
\end{equation}

Finally, the MSE loss is added as a regularizer term and the entire network is optimized through back-propagation to minimize the total loss, which is defined as:
\begin{equation}
\mathcal{L}=\mathcal{L}_{ce}^{f} + \mathcal{L}_{ce}^{d}+\gamma\mathcal{L}_{mse},
\end{equation}
where $\gamma$ represents a hyperparameter that controls the influence of the MSE loss term in relation to the cross-entropy losses.
\section{Experiments and Analysis}
\subsection{Datasets and Implementation Details}
To enhance the data scale for copy-move forgery, Wu et al. \cite{wu2018busternet} introduced the USCISI dataset, which was compiled from two existing datasets: the MIT SUN2012 dataset \cite{xiao2010sun} and the Microsoft COCO dataset \cite{lin2014microsoft}. The USCISI dataset comprises over 100,000 image samples, ranging in size from 340 $\times$ 260 to 1068 $\times$ 994 pixels. Following the experimental methodology outlined in \cite{wu2018busternet}, 80,000 samples were utilized for training purposes, while the remaining samples were divided equally for validation and testing. The generalization ability of the proposed network was assessed by employing the CoMoFoD dataset \cite{tralic2013comofod}, a representative copy-move dataset. CoMoFoD contains a total of 5,000 images, including 200 base forged images and 25 categories of attacks.

The proposed method was implemented using the PyTorch framework and trained and evaluated on an NVIDIA A100 graphics processing unit. During training, we utilized the Adam optimizer  \cite{kingma2014adam} with an initial learning rate of 0.001 and a weight decay of 0.0005. To adapt the learning rate during training, we employed the ``poly" learning rate decay policy. This policy adjusts the initial learning rate by multiplying it with $(1-\text{iter}/\text{maxiter})^{\text{power}}$, where $\text{power}$ is set to 0.9 after each iteration. The training process was terminated after 30 epochs, and a batch size of 64 was used for training. After each epoch, we evaluated the model's performance on the validation set. The best-performing model, determined by the highest mean $F_1$ score, was selected for evaluation on the test set.

\subsection{Experimental Result}
To evaluate the performance of our proposed method, we conducted a rigorous comparison with established forgery detection and source/target distinguishment techniques, namely BusterNet \cite{wu2018busternet} and DOA-GAN \cite{islam2020doa}.

\begin{figure}[!t]
\centering
\includegraphics[width=\linewidth]{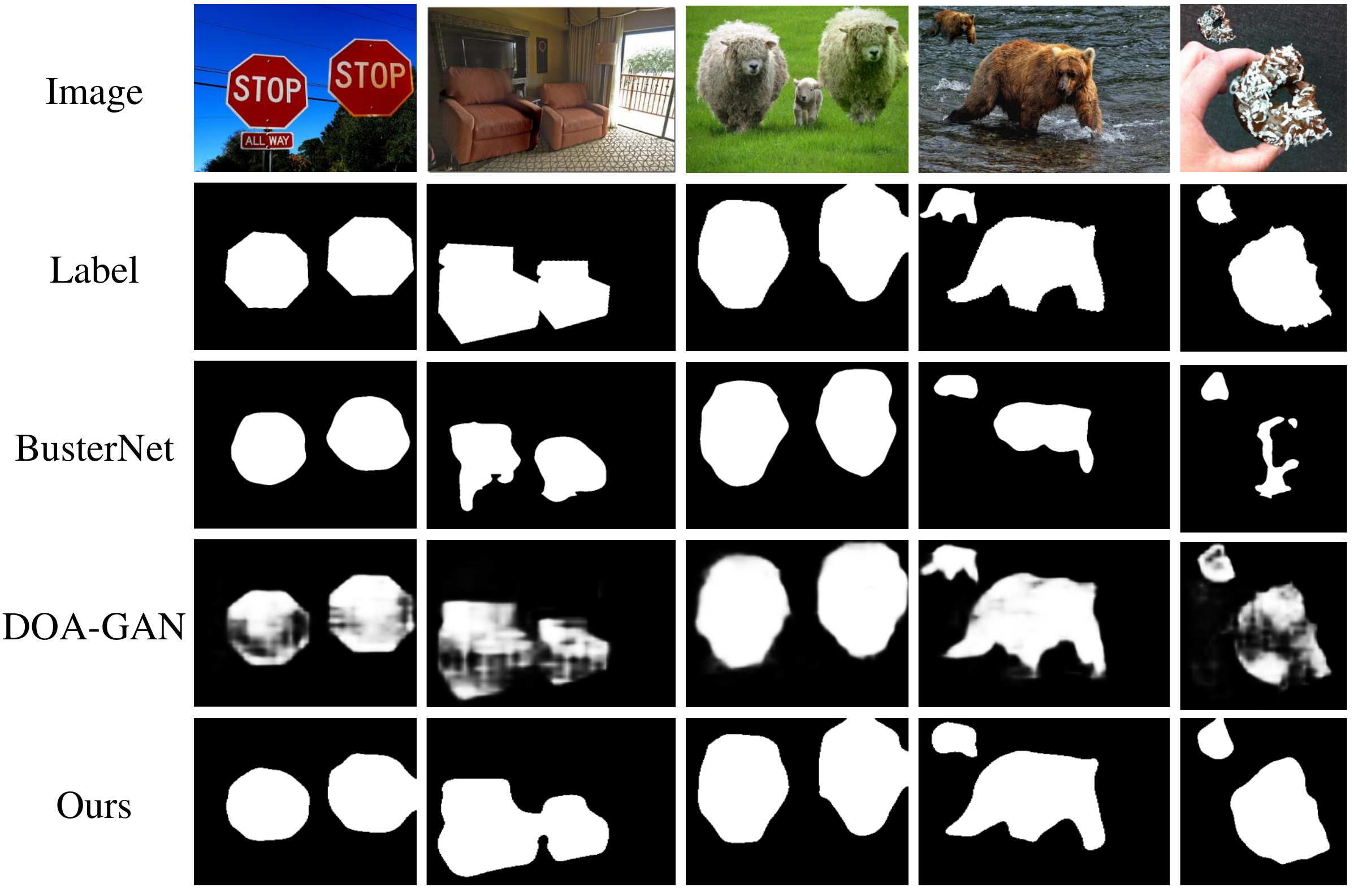}
\caption{Visualized examples of the copy-move forgery detection maps on the USCISI dataset.}
\label{fig_1}
\end{figure}

\begin{figure}[!t]
\centering
\includegraphics[width=0.8\linewidth]{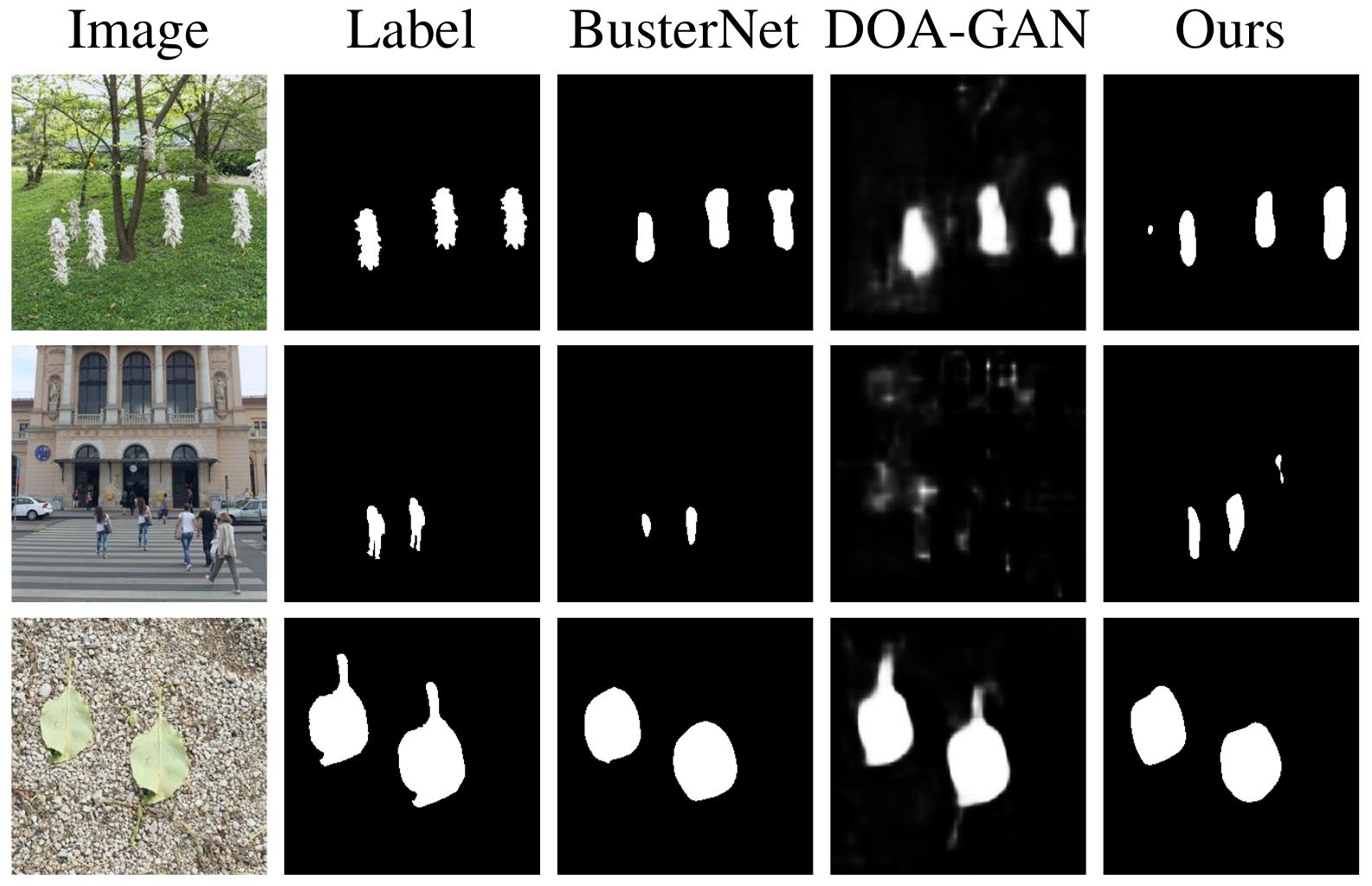}
\caption{Visualized examples of the copy-move forgery detection maps on the CoMoFoD dataset.}
\label{fig_4}
\end{figure}

For the task of copy-move forgery detection, all manipulated regions within the images were identified as forged regions. We employed a meticulous evaluation approach, measuring pixel-level precision, recall, and F1-score for each image. The average scores across the entire USCISI dataset were reported. The detection accuracy, along with the visualized results, are presented in Table \ref{tab1} and Fig. \ref{fig_1}. Considering the CoMoFoD dataset, which incorporates diverse postprocessing operations applied to the base images to evaluate the robustness of the methods, we compute the mean F1-score within each category. For the purpose of correctly identifying detected images, we define F1-scores exceeding 0.5 as successful detection. This criterion is detailed in Tables \ref{tab2}-\ref{tab3}. Additionally, we present a subset of visual examples in Fig. \ref{fig_4}. It can be observed that the proposed research outperforms other competitive methods in accurately extracting forged region.

\begin{table}
\centering
\caption{Binary Detection Performances on the USCISI Dataset Expressed in Terms of Pixel-Level Evaluation}
\begin{tabular}{c|ccc}
\toprule
\toprule[0.5pt]
Algorithm & Precision  & Recall & F1 \\ \midrule
BusterNet & 55.61 & 57.74 & 52.16  \\ 
DOA-GAN  & 78.09 & 55.95 & 60.80 \\ 
Ours  & \textbf{69.89} & \textbf{65.90} & \textbf{65.55} \\
\bottomrule[0.5pt] \bottomrule
\end{tabular}
\label{tab1}
\end{table}

\begin{table*}
\centering
\caption{Number of Correctly Detected Images on the CoMoFoD Dataset Under Different Attacks}
\resizebox{\textwidth}{!}{
\begin{tabular}{c|ccccccccccccccccc}
\toprule
\toprule[0.5pt]
Algorithm & BASE  & BC1 & BC2 & BC3 & CA1 & CA2 & CA3 & CR1 & CR2 & CA3 & IB1 & IB2 & IB3 & NA1 & NA2 & NA3 \\ \midrule
BusterNet & 120 & 119 & 115 & 109 & 119 & 119 & 115 & 120 & 120 & 119 & 177 & 99 & 94 & 101 & 105 & 118 \\ 
DOA-GAN  & 89 & 89 & 83 & 78 & 89 & 90 & 92 & 89 & 90 & 88 & 91 & 86 & 65 & 55 & 72 & 84 \\ 
Ours  & \textbf{159} & \textbf{162} & \textbf{167} & \textbf{178} & \textbf{150} & \textbf{158} & \textbf{152} & \textbf{157} & \textbf{156} & \textbf{158} & \textbf{161} & \textbf{160} & \textbf{152} & \textbf{169} & \textbf{168} & \textbf{157} \\
\bottomrule[0.5pt] \bottomrule
\end{tabular}
\label{tab2}}
\end{table*}

\begin{table}
\centering
\caption{Number of Correctly Detected Images on the CoMoFoD Dataset Under Different Attacks}
\resizebox{0.5\textwidth}{!}{
\begin{tabular}{c|cccccccccc}
\toprule
\toprule[0.5pt]
Algorithm & JC1 & JC2 & JC3 & JC4 & JC5 & JC6 & JC7 & JC8 & JC9 \\ \midrule
BusterNet & 55 & 71 & 84 & 96 & 101 & 104 & 111 & 110 & 106  \\ 
DOA-GAN  & 21 & 38 & 53 & 67 & 76 & 83 & 87 & 94 & 86  \\ 
Ours  & \textbf{159} & \textbf{141} & \textbf{162} & \textbf{149} & \textbf{155} & \textbf{157} & \textbf{148} & \textbf{172} & \textbf{166} \\
\bottomrule[0.5pt] \bottomrule
\end{tabular}
\label{tab3}}
\end{table}

\begin{figure}[!t]
\centering
\includegraphics[width=\linewidth]{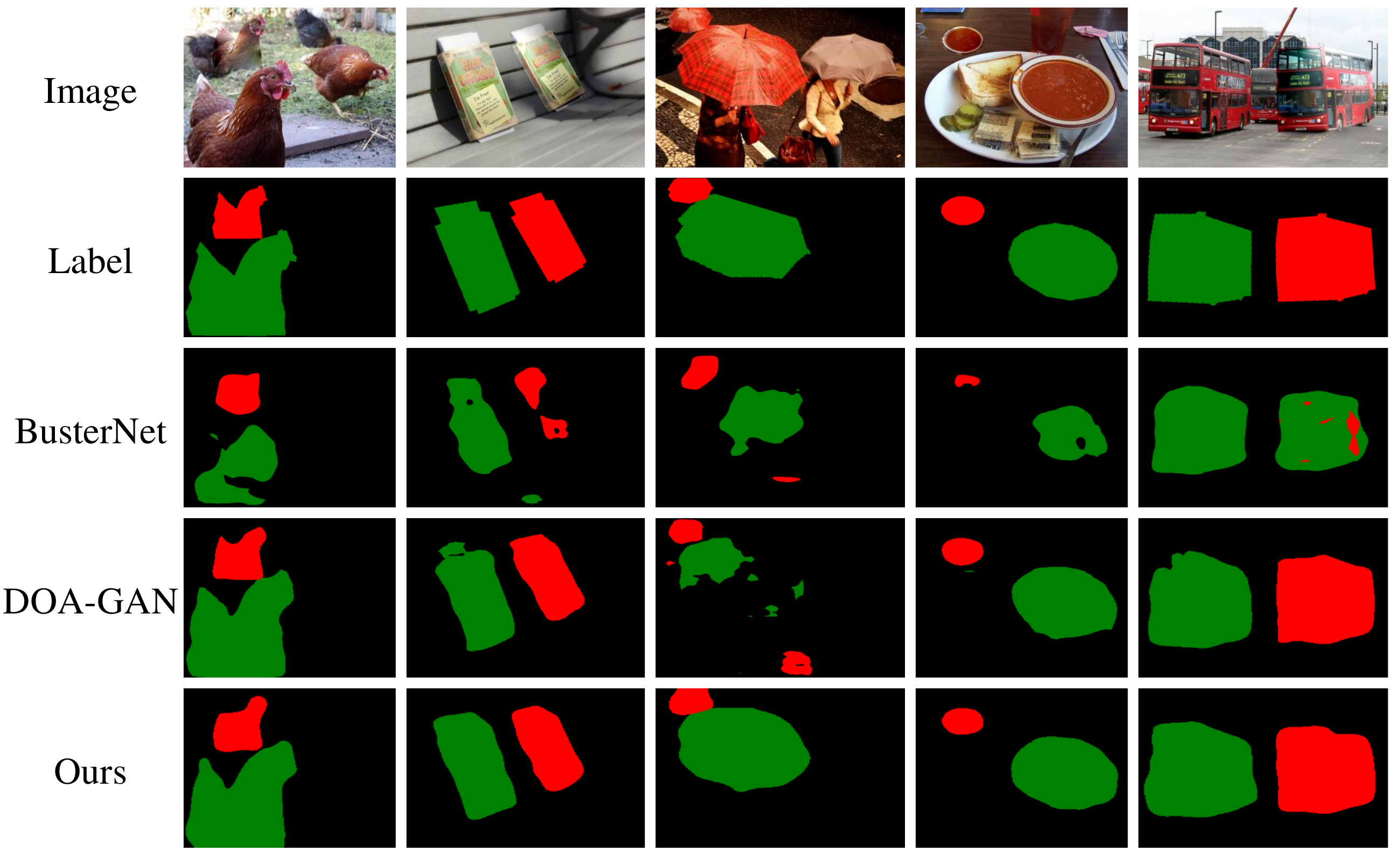}
\caption{Visualized examples of the source/target distinguishment maps on the USCISI dataset.}
\label{fig_3}
\end{figure}

\begin{figure}[!t]
\centering
\includegraphics[width=0.8\linewidth]{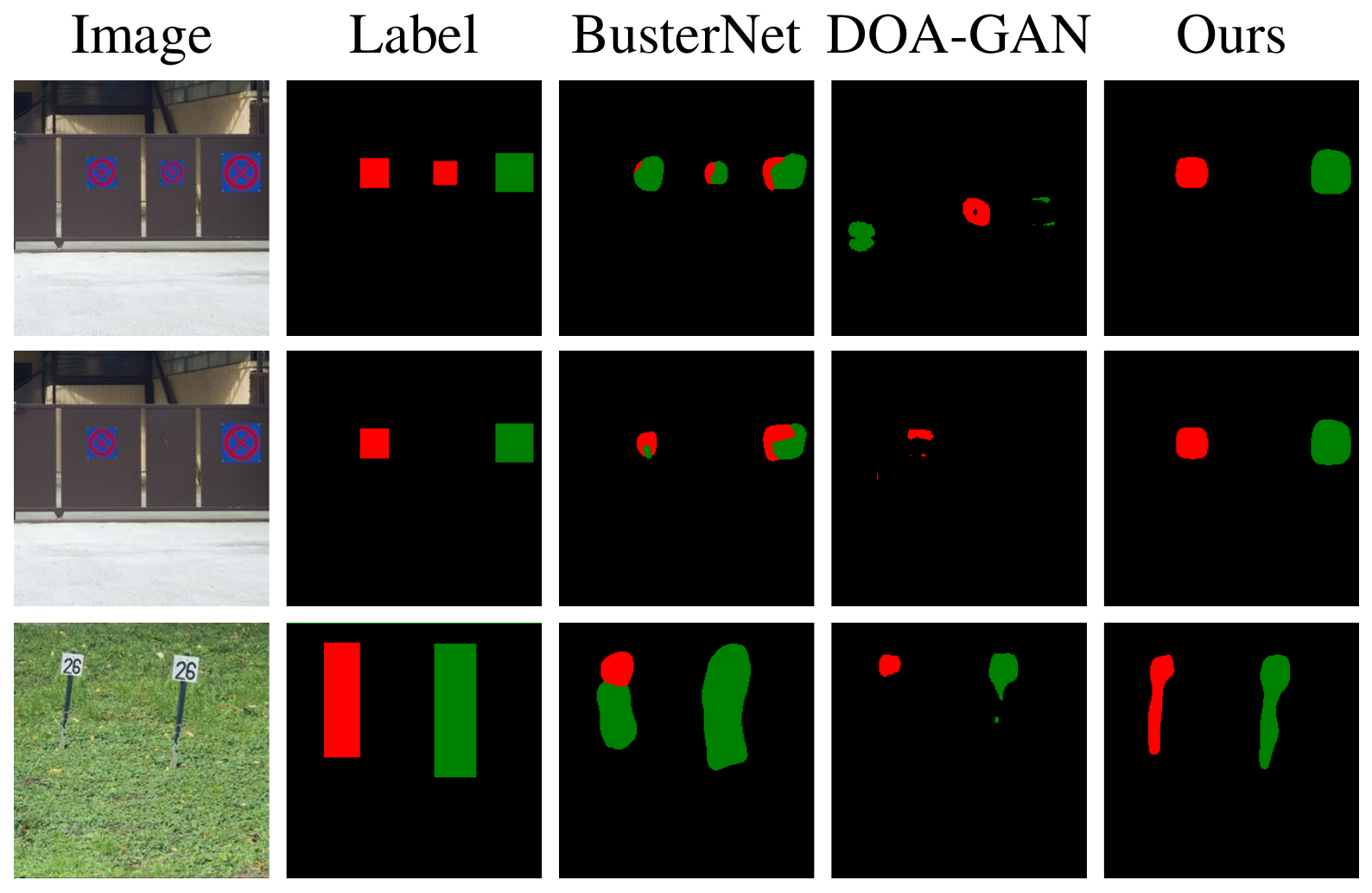}
\caption{Visualized examples of the source/target distinguishment maps on the CoMoFoD dataset.}
\label{fig_2}
\end{figure}

The quantitative results and visualization examples of the source/target distinguish task are presented in Table \ref{tab4} and Figs. \ref{fig_3}-\ref{fig_2}. It is important to note that the model was trained and validated specifically on the USCISI dataset. Among the evaluated models, DOA-GAN demonstrated the highest performance on the original data source, while BusterNet exhibited the best transferability across different sources and targets.

Although our network achieved the second-best performance compared to BusterNet and DOA-GAN on both datasets, it is worth mentioning that the binary detection task's optimal performance does not necessarily translate to the best results for the source/target differentiate task. This observation highlights the importance of considering specific task requirements when evaluating model performance.

Furthermore, it is crucial to explore the transferability of the copy-move detection network across various datasets. While BusterNet demonstrated superior transferability in this study, the relatively low scores in source and target classes illustrate its robustness in locating the manipulated regions still needs improvements. Continued research in this area will provide valuable insights into the generalization capabilities of copy-move detection networks and their effectiveness in different contexts.

\begin{table}
\centering
\caption{Source/Target Distinguishment Performances on the USCISI and the CoMoFoD Datasets Expressed in Terms of Pixel-Level Evaluation}
\begin{tabular}{ccc|ccc}
\toprule
\toprule[0.5pt]
\multicolumn{1}{c}{}     & \multicolumn{1}{c}{}      & \multicolumn{1}{c|}{} & BusterNet & DOA-GAN & Ours  \\ \hline
\multirow{9}{*}{USCISI}  & \multirow{3}{*}{source}   & precision            & 37.45     & \textbf{76.18}   & 63.76 \\ 
                         &                           & recall               & 16.71     & \textbf{63.62}   & 61.21 \\
                         &                           & F1                   & 20.75     & \textbf{66.46}   & 60.83 \\ \cline{2-6}
                         & \multirow{3}{*}{target}   & precision            & 48.69     & \textbf{85.14}   & 74.89 \\
                         &                           & recall               & 21.75     & \textbf{80.03}   & 70.45 \\
                         &                           & F1                   & 26.84     & \textbf{81.28}   & 70.92 \\ \cline{2-6}
                         & \multirow{3}{*}{pristine} & precision            & 92.36     & \textbf{96.97}   & 96.72 \\
                         &                           & recall               & 99.25     & \textbf{98.85}   & 97.64 \\
                         &                           & F1                   & 95.55     & \textbf{97.87}   & 97.13 \\ \hline
\multirow{9}{*}{CoMoFoD} & \multirow{3}{*}{source}   & precision            & \textbf{21.53}     & 12.64   & 10.70 \\
                         &                           & recall               & 6.98      & 4.45    & \textbf{12.40} \\
                         &                           & F1                   & \textbf{8.86}      & 5.34    & 7.89  \\ \cline{2-6}
                         & \multirow{3}{*}{target}   & precision            & \textbf{33.11}     & 8.28    & 16.65 \\
                         &                           & recall               & \textbf{28.93}     & 2.83    & 12.39 \\
                         &                           & F1                   & \textbf{27.26}     & 3.42    & 10.57 \\ \cline{2-6} 
                         & \multirow{3}{*}{pristine} & precision            & \textbf{97.75}     & 94.76   & 95.21 \\
                         &                           & recall               & 99.45     & \textbf{99.63}   & 98.02 \\
                         &                           & F1                   & \textbf{98.58}     & 97.00   & 96.41 \\
\bottomrule[0.5pt] \bottomrule
\end{tabular}
\label{tab4}
\end{table}

\subsection{Ablation Study}
To comprehensively evaluate the effectiveness of the Mean Squared Error (MSE) loss and the transformer-based encoder, an ablation study was carried out on the USCISI dataset. The objective of this study was to investigate the impact of these components on the overall performance of the proposed network. The results of this study are presented in Table \ref{tab5}, which showcases the performance of the network across two tasks. It is evident from the table that both the MSE loss and the transformer contribute to enhancing the segmentation performance. The utilization of the MSE loss aids in minimizing the discrepancy between the predicted and ground truth segmentation maps, facilitating accurate boundary delineation. The transformer-based encoder, with its ability to capture long-range dependencies and contextual information, enriches the network's understanding of spatial relationships and structural patterns within the dataset. As a result, it effectively improves the overall segmentation accuracy and robustness of the system.

\begin{table*}
\centering
\caption{Ablation Study of the Proposed Network on the USCISI Dataset.}
\resizebox{\textwidth}{!}{
\begin{tabular}{cc|ccc|ccc|ccc|ccc|ccc}
\toprule
\toprule[0.5pt]
\multirow{3}{*}{MSE loss} & \multirow{3}{*}{Transformer} & \multicolumn{6}{c|}{Copy-Move Detection}                                         & \multicolumn{9}{c}{Source/Target Distinguishment} \\
&& \multicolumn{3}{c|}{Source \& Target} & \multicolumn{3}{c|}{Pristine}          & \multicolumn{3}{c|}{Source}    & \multicolumn{3}{c|}{Target}    & \multicolumn{3}{c}{Pristine}  \\ 
&& Precision   & Recall     & F1   & Precision & Recall & F1 & Precision & Recall & F1 & Precision & Recall & F1 & Precision & Recall & F1 \\ \midrule
 & & 63.07   & 48.07   & 51.30 & 94.66 & 97.29 & 95.87 & 47.93 & 38.48 & 40.10 & 71.31 & 66.94 & 66.59 & 94.63 & 97.34 & 95.88\\ 
\checkmark & & 61.93   & 50.94   & 52.97 & 94.91 & 96.93 & 95.82 & 48.49 & 42.77 & 43.02 & 73.00 & 66.89 & 67.32 & 94.90 & 96.92 & 95.81 \\ 
 & \checkmark  & 67.56   & 63.39   & 62.68 & 96.44 & 97.10 & 96.71 & 57.77 & 56.49 & 55.16 & 72.28 & 76.76 & 72.73 & 96.43 & \textbf{97.13} & 96.72 \\ 
\checkmark & \checkmark  & \textbf{67.83}   & \textbf{65.08}   & \textbf{64.13} & \textbf{96.65} & \textbf{97.09} & \textbf{96.82} & \textbf{58.90} & \textbf{58.50} & \textbf{56.91} & \textbf{74.34} & \textbf{77.49} & \textbf{74.29} & \textbf{96.65} & 97.12 & \textbf{96.83}  \\ 
\bottomrule[0.5pt] \bottomrule
\end{tabular}}
\label{tab5}
\end{table*}

\subsection{Parametric Analysis}
In this subsection, we delve into examining the impact of the value of $\gamma$ and the depths of the transformer encoder on network performance. Our investigation, as presented in Table \ref{tab6}, reveals that the detection and distinguishment performance undergo significant improvement with higher values of $\gamma$. Specifically, we find that the suggested value of $\gamma$ for optimal results is $1000$. Moreover, the depths of the transformer encoder also play a crucial role in network performance. Based on the findings in Table \ref{tab7}, it is recommended to employ a depth of either $1$ or $2$ for the transformer encoder in order to achieve desirable outcomes.
\begin{table*}
\centering
\caption{Parametric Analysis About the Value of $\gamma$.}
\resizebox{\textwidth}{!}{
\begin{tabular}{c|ccc|ccc|ccc|ccc|ccc}
\toprule
\toprule[0.5pt]
\multirow{3}{*}{$\gamma$} & \multicolumn{6}{c|}{Copy-Move Detection}                                         & \multicolumn{9}{c}{Source/Target Distinguishment} \\
& \multicolumn{3}{c|}{Source \& Target} & \multicolumn{3}{c|}{Pristine}          & \multicolumn{3}{c|}{Source}    & \multicolumn{3}{c|}{Target}    & \multicolumn{3}{c}{Pristine}  \\ 
& Precision   & Recall     & F1   & Precision & Recall & F1 & Precision & Recall & F1 & Precision & Recall & F1 & Precision & Recall & F1 \\ \midrule
0.01  & 65.36   & 63.38   & 61.37 & 96.47 & 96.75 & 96.54 & 56.07 & 56.91 & 54.30 & 72.03 & 73.91 & 71.01 & 96.42 & 96.84 & 96.57 \\ 
0.1 & 68.07   & 62.66   & 62.36 & 96.41 & 97.19 & 96.74 & 57.64 & 55.47 & 54.32 & 72.95 & 76.49 & 73.04 & 96.40 & 97.21 & 96.74 \\ 
1  & 67.83   & 65.08   & 64.13 & 96.65 & 97.09 & 96.82 & 58.90 & 58.50 & 56.91 & 74.34 & \textbf{77.49} & 74.29 & 96.65 & 97.12 & 96.83 \\ 
10  & 68.34  & 63.20  & 63.06 & 96.49 & 97.24 & 96.81 & 58.67 & 56.21 & 55.45 & 75.46 & 74.91 & 73.61 & 96.43 & 97.35 & 96.83 \\ 
100 & 69.76 & 64.01 & 64.25 & 96.59 & \textbf{97.42} & 96.96 & 60.17 & 57.05 & 56.77 & 75.88 & 74.96 & 73.80 & 96.51 & \textbf{97.58} & 96.99 \\
1000 & \textbf{69.89} & \textbf{65.90} & \textbf{65.55} & \textbf{96.75} & 97.34 & \textbf{97.00} & \textbf{61.10} & \textbf{59.74} & \textbf{58.64} & \textbf{76.00} & 76.83 & \textbf{74.98} & \textbf{96.73} & 97.38 & \textbf{97.01} \\
\bottomrule[0.5pt] \bottomrule
\end{tabular}}
\label{tab6}
\end{table*}

\begin{table*}
\centering
\caption{Parametric Analysis About the Depths of Two Vision Transformer.}
\resizebox{\textwidth}{!}{
\begin{tabular}{c|ccc|ccc|ccc|ccc|ccc}
\toprule
\toprule[0.5pt]
\multirow{3}{*}{Depth} & \multicolumn{6}{c|}{Copy-Move Detection}                                         & \multicolumn{9}{c}{Source/Target Distinguishment} \\
& \multicolumn{3}{c|}{Source \& Target} & \multicolumn{3}{c|}{Pristine}          & \multicolumn{3}{c|}{Source}    & \multicolumn{3}{c|}{Target}    & \multicolumn{3}{c}{Pristine}  \\ 
& Precision   & Recall     & F1   & Precision & Recall & F1 & Precision & Recall & F1 & Precision & Recall & F1 & Precision & Recall & F1 \\ \midrule
1 & 69.89 & \textbf{65.90} & \textbf{65.55} & \textbf{96.75} & 97.34 & 97.00 & 61.10 & 59.74 & 58.64 & \textbf{76.00} & \textbf{76.83} & \textbf{74.98} & \textbf{96.73} & 97.38 & 97.01 \\ 
2 & \textbf{70.15}   & 65.33   & 65.52 & 96.18 & \textbf{97.58} & \textbf{97.12} & \textbf{63.76} & \textbf{61.21} & \textbf{60.83} & 74.89 & 70.45 & 70.92 & 96.72 & \textbf{97.64} & \textbf{97.13} \\ 
3  & 59.79 & 57.56  & 56.21 & 95.82 & 96.11 & 95.89 & 50.62 & 50.16 & 48.80 & 65.94 & 68.43 & 64.91 & 95.80 & 96.15 & 95.90 \\ 
4  & 57.94 & 44.32  & 45.49 & 94.33 & 96.58 & 95.31 & 40.56 & 39.60 & 36.86 & 57.92 & 58.31 & 54.61 & 94.74 & 95.57 & 95.00 \\ 
\bottomrule[0.5pt] \bottomrule
\end{tabular}}
\label{tab7}
\end{table*}

\section{Conclusion}
In this study, we aim to present an approach for enhancing forgery detection and source/target distinguishment using an end-to-end network. Our proposed method simultaneously generates a forgery detection map and a source/target distinguishment map. To achieve this, we employ a transformer-based encoder comprising two vision transformers. Unlike directly memorizing the entire dataset, the proposed approach focuses on capturing the copy and moved regions by emphasizing the inner features associated with the manipulated areas. To strike a balance between detection accuracy and localization, we introduce the Mean Squared Error (MSE) similarity between the two maps. This similarity measure is calculated and utilized in the back-propagation process to minimize the overall training network. By leveraging this strategy, we aim to improve the network's ability to accurately identify and locate manipulated regions. Although the experimental results demonstrate that the proposed toy end-to-end network surpasses the performance of existing methods in the binary detection task, there is still room for further exploration in developing reliable and robust model that can effectively disambiguate changed regions and perform consistently on other datasets.

\bibliographystyle{IEEEtran}
\bibliography{ref}

\vfill

\end{document}